\documentclass[sigconf]{acmart}
\AtBeginDocument{%
  }

\setcopyright{none}
\acmISBN{978-1-4503-XXXX-X/2025/08}

\usepackage{multirow}
\usepackage{makecell}
\usepackage{bbm, dsfont}

\begin{document}

\title{Resolution-Aware Retrieval Augmented Zero-Shot Forecasting}

\author{Iman Deznabi}
\email{iman@cs.umass.edu}
\affiliation{%
  \department{Manning College of Information and Computer Sciences}
  \institution{University of Massachusetts Amherst}
  \city{Amherst}
  \state{MA}
  \country{USA}
}

\author{Peeyush Kumar}
\email{peeyush.kumar@microsoft.com}
\affiliation{%
  \institution{Microsoft Research}
  \city{Redmond}
  \state{WA}
  \country{USA}
}

\author{Madalina Fiterau}
\email{mfiterau@cs.umass.edu}
\affiliation{%
  \department{Manning College of Information and Computer Sciences}
  \institution{University of Massachusetts Amherst}
  \city{Amherst}
  \state{MA}
  \country{USA}
}

\renewcommand{\shortauthors}{Deznabi et al.}

\begin{abstract}
Zero-shot forecasting aims to predict outcomes for previously unseen conditions without direct historical data, posing a significant challenge for traditional forecasting methods. We introduce a Resolution-Aware Retrieval-Augmented Forecasting model that enhances predictive accuracy by leveraging spatial correlations and temporal frequency characteristics. By decomposing signals into different frequency components, our model employs resolution-aware retrieval, where lower-frequency components rely on broader spatial context, while higher-frequency components focus on local influences. This allows the model to dynamically retrieve relevant data and adapt to new locations with minimal historical context.

Applied to microclimate forecasting, our model significantly outperforms traditional forecasting methods, numerical weather prediction models, and modern foundation time series models, achieving 71\% lower MSE than HRRR and 34\% lower MSE than Chronos on the ERA5 dataset.
Our results highlight the effectiveness of retrieval-augmented and resolution-aware strategies, offering a scalable and data-efficient solution for zero-shot forecasting in microclimate modeling and beyond.
\end{abstract}



\maketitle

\section{Introduction}
Zero-shot forecasting aims to predict outcomes for previously unseen conditions without direct historical data. This is especially important in scenarios where collecting context-specific data is costly or infeasible. Our approach addresses this challenge by leveraging spatial correlations and temporal frequency characteristics, enabling effective forecasting through knowledge transfer from similar contexts.
We will use the problem of microclimate prediction as the primary application, however, the models can be used in different spatiotemporal problems and zero-shot forecasting scenarios.
We will introduce models that can bring personalization at different resolutions leveraging spatial correlations and temporal frequency characteristics, enabling effective zero-shot forecasting by incorporating knowledge from similar contexts.

Microclimate prediction involves the forecasting and analysis of localized variations in weather conditions within specific, relatively small regions. Unlike broader regional or macro-climate predictions, which provide generalized weather information for large areas, microclimate predictions focus on understanding the intricacies of weather patterns within smaller, more homogeneous areas. 

Previous works \cite{kumar2021micro, ajani2023greenhouse, eleftheriou2018micro, moonen2012urban} have highlighted the importance of microclimate prediction. For instance, \cite{kumar2021micro} discusses a farmer whose decision to fertilize fields was based on data from a weather station 50 miles away, leading to significant crop damage due to localized temperature variations. This situation underscores the crucial role of accurate microclimate prediction. Apart from agriculture, microclimate prediction is indispensable in many fields such as forestry, architecture, urban planning, ecology conservation, and maritime activities and it plays a critical role in optimizing decision-making and resource allocations. It empowers stakeholders to make informed decisions and adapt to localized climate variations effectively.

Let's consider predicting climate variables for a new farm or location, where we lack historical data. Traditional methods for time series forecasting including ones developed for microclimate prediction often require extensive data collection, including ground-based sensors or weather stations, which can be costly and limited in coverage. To solve this problem we look at zero-shot microclimate forecasting which refers to the task of predicting fine-grained environmental conditions at specific locations without relying on direct observations or measurements.

In this work, we developed a microclimate prediction deep learning model based on the concept of transfer and retrieval-based learning, where knowledge acquired from several locations is used to make predictions in another location with limited or no available data.
Transfer learning enables the application of predictive models trained on existing climate data to estimate microclimate variables such as temperature, in previously unmonitored areas.

While large foundation time-series models \cite{ansari2024chronos, liang2024foundation, das2023decoder, goswami2024moment, nie2022time} exhibit strong generalization capabilities, they struggle to adapt to new, unseen locations due to the lack of specific contextual data. Existing deep learning models for spatiotemporal weather modeling \cite{grigsby2021long, bojesomo2023novel, pathak2022fourcastnet} are often computationally expensive and limited to lower spatial resolutions or global scales. Additionally, these models typically assume data points are on a grid with equal spacing, which does not hold in many microclimate scenarios. We address these issues by developing models that can transfer data using Attention mechanism \cite{bahdanau2014neural}
to handle arbitrary distances between points. Numerical Weather Prediction (NWP) models \cite{coiffier2011fundamentals, kimura2002numerical, lorenc1986analysis} and adaptive learning methods \cite{ragab2022self, lu2023multi, zhang2015multi} face similar limitations. Our model extends recent retrieval-augmented forecasting methods \cite{jing2022retrieval, deznabi2024zero, liu2024retrieval} by incorporating resolution-aware retrieval and attention-based adaptive learning, offering a novel approach to zero-shot forecasting.

Adaptive and retrieval augmented learning for time series forecasting has gained significant attention in recent years, with various methodologies being explored to enhance forecasting accuracy. We will use this technique as a way of personalization in time series, by bringing similar data to the target stream of interest, this technique has been extensively used in large language models (LLMs)\cite{salemi2023lamp, salemi2024optimization, mathur2023personalm}. 
Very recently retrieval augmented forecasting models are introduced for time-series data \cite{jing2022retrieval, tire2024retrieval, liu2024retrieval, yang2024timerag, zhang2024timeraf}, One notable approach for time series is ReTime\cite{jing2022retrieval}, which proposes a two-stage method, consisting of a relational retrieval and a content synthesis using a content attention layer, and theoretically and empirically show that this decreases uncertainty in forecasting. Additionally, Memory Augmented Graph Learning Networks (MAGL) have been proposed to capture global historical features of multivariate time series, outperforming traditional methods by leveraging both intra-series temporal and inter-series spatial correlations \cite{liu2022memory}. However, these models rely on extensive prior context at the time of forecasting, whereas our approach assumes that the available prior context for the target is highly limited.

Signal decomposition methods, particularly wavelet-based approaches, have shown great promise for modeling non-stationary time series by capturing both local and global temporal dynamics. Wavelet transforms enable multiscale analysis, making them especially effective in scenarios with varying frequency behavior across time and space \cite{rhif2019wavelet, zhu2023time}. These concepts have been extended to deep learning models for time series \cite{gupta2021multiwavelet, xiao2022coupled, zhang2022first, masserano2024enhancing}. These works illustrate the importance of isolating temporal dynamics across frequency bands. Inspired by this, our model introduces a resolution-aware retrieval mechanism, which decomposes input signals into different frequency components and retrieves context adaptively—using distant stations to inform low-frequency components and nearby stations for high-frequency variations.

Our model also draws on domain adaptation methods \cite{wilson2020multi, ott2022domain, jin2022domain, xiang2023two, he2023domain} to design the transfer component that learns to map knowledge from locations with abundant data to new, unobserved locations. Using attention, we learn personalized embeddings for the target location. We further enhance this by using frequency decomposition to assign different sets of retrieved contexts per frequency band, reflecting the intuition that spatial influence varies across temporal scales.


In summary, the contributions of this work are:
\begin{itemize}
    \item We developed a model architecture that uses attention for retrieval augmented forecasting of microclimate conditions where little or no historical data is available for the target.
    \item We developed a resolution aware retrieval method which retrieves a bigger set of points for the lower frequencies of the data
    \item By putting the resolution aware retrieval method and retrieval augmented forecasting model together, we propose a framework that is able to accurately forecast microclimate parameters for a location that has little or no available historical data.
    \item We trained and evaluated these models on real-world data and make our models available as a foundational model for zero-shot microclimate prediction.
\end{itemize}

\section{Methodology}
\begin{figure*}[t]
    \centering
    \includegraphics[width=0.8\textwidth]{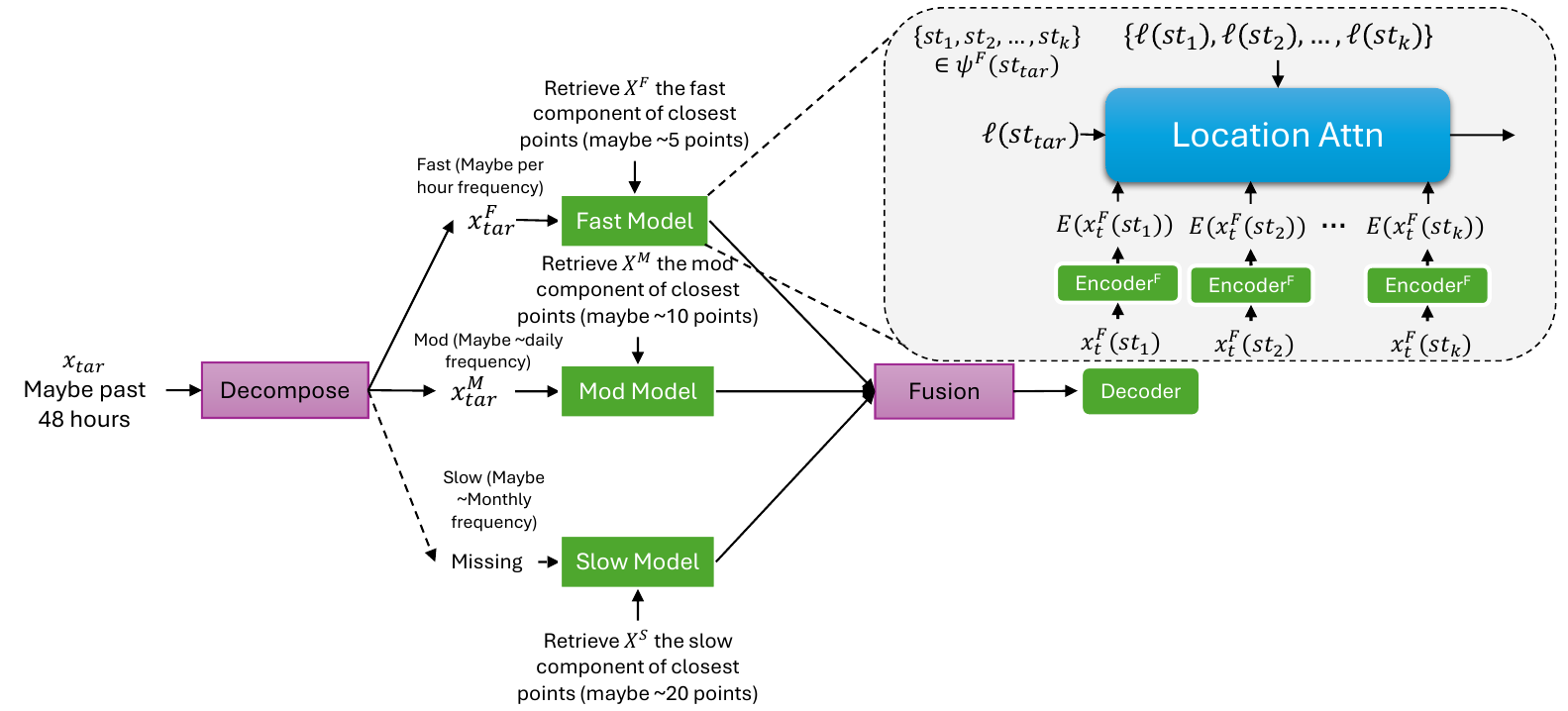}
    \vspace{-1em}
    \caption{Overall structure of the model with location attention transfer component and resolution-aware retrieval. The limited available context for the target location ($x_{tar}$) is first decomposed into 3 levels of fast, moderate and slow, for each of these components, our retrieval augmented forecasting model with location attention is used. For the fast components we retrieve the fast components of a smaller set of closer points to the target, while for slower components we retrieve more points.}
    \label{fig:GNN_trans_model}
    \vspace{-0.5em}
\end{figure*}
The overall model structure is shown in Figure~\ref{fig:GNN_trans_model}, for a target location we have limited past data available, assumed here as 48 hours, we decompose this past data into let's say fast, moderate, and slow frequencies using wavelet decomposition or other decomposition techniques, then for each frequency we gather a different set of points, that increase in size for slower frequencies. For each set we use our retrieval augmented model to get the embedding for that corresponding frequency and then fuse these embeddings and pass them to the decoder to get the forecast values. The fusion function is the inverse of wavelet decomposition. 
Note that for the target location, due to the limited availability of context data (as illustrated in Figure~\ref{fig:GNN_trans_model}), some components might not be accessible. Consequently, our model relies solely on the retrieved set to make predictions for those specific frequencies. In this section, we will first define the problem and then discuss the structure of the retrieval-augmented model, the proposed retrieval techniques, the location embedding methods used for retrieval and within the transfer module, and finally, the transfer component within the retrieval-augmented forecasting model.
\subsection{Problem Formulation}
In this study, we aim to forecast climate parameters over a time horizon $L_y$ starting from the current time $t$. Our task involves predicting climate parameter values $\mathcal{Y}_t = \{y_{t+1}, y_{t+2}, ..., y_{t+L_y} | y_i \in \mathbb{R}\}$. To achieve this, we take as input a limited preceding window (with size $L_x$) of relevant climate parameters, denoted as $\mathcal{X}_t = \{x_{t-L_x}, x_{t-L_{x}+1}, ..., x_{t} | x_i \in \mathbb{R}^n\}$ where $x_t$ represents the $n$ available climate parameters for input at time $t$, and $y_{t'}$ denotes the target climate parameters of interest at time $t'$. Our predictions are specific to a particular location, referred to as the ``target station'' ($st_{tar}$), characterized by geographic data encompassing latitude, longitude, and elevation, denoted as $l(st_{tar})$. We will use the notations $\mathcal{X}_t(st_i)$, $\mathcal{Y}_t(st_i)$ and $\hat{\mathcal{Y}}_t(st_i)$ to denote relevant past climate parameters, target climate parameter and the forecasts values for station $st_i$ at time $t$. The historical dataset ($\mathcal{H}$) consists of past climate parameter values measured at time intervals preceding $t$ for multiple stations $\mathcal{H} = \{(\mathcal{X}_{t'}, \mathcal{Y}_{t'}) | t' < t\}$, forming the foundation for our predictive modeling. In the zero-shot scenario, historical data within $\mathcal{H}$ does not include any information about the target station ($(\mathcal{X}_t(st_{tar}), \mathcal{Y}_t(st_{tar})) \notin \mathcal{H} \; \forall t$). Consequently, we rely solely on available input data from other sources to forecast climate parameters at $st_{tar}$ and the small immediate preceding window of size $L_x$ for the target station ($L_x$ can be zero for cold start forecasting). This method addresses this zero-shot prediction problem by developing and evaluating models capable of accurate climate forecasting at $st_{tar}$ in the absence of historical data specific to that location. 
  
\subsection{Retrieval-Augmented Forecasting Model}
In the zero-shot forecasting setting, the past values (or context) available for the target point are limited. Therefore, to improve forecasting accuracy, it is crucial to leverage information from other reference points. Our model addresses this by learning the following probability distribution which we call \textit{retrieval-augmented forecasting model}:
$$P(\mathcal{Y}_t(st_{tar})|\mathcal{X}_t(st_{tar}), \mathcal{X}_t(st^{R_{tar}}_1), \mathcal{X}_t(st^{R_{tar}}_2), ...)$$ 
where $\{\text{st}^{R_{tar}}_1, \text{st}^{R_{tar}}_2, ...\} \in \psi(\text{st}_{tar})$ represents the set of reference points for the target point. The function $\psi(\text{st}_i)$ retrieves these reference points based on a distance function $d(\text{st}_i, \text{st}_j)$: $\psi(\text{st}_i) = \text{argmin}^{\text{topk}}_{\text{st}_j} \, d(\text{st}_i, \text{st}_j)$. We will give more details about the choice of the distance function $d$ and the retrieval method $\psi$ in section~\ref{chapter5-retrieval}.

For simplification, we denote the past values of the retrieved reference points as $\mathcal{X}^{R}_t$. The final forecast for the target point is then given by: $$\hat{\mathcal{Y}}_t(st_{tar}) = \phi(\mathcal{X}_t, \mathcal{X}^{R}_t)$$ 
where $\phi$ denotes the retrieval-augmented forecasting model.


The central concept behind this architecture is to develop a ``transfer'' component capable of extrapolating knowledge from the retrieved stations $\psi(st_{tar})$ which we possess training data for based on their location. This function then transfers this knowledge into the encoding of a target station, even if we lack precise training data for that particular station. Subsequently, we employ the decoder to generate forecast for the target station using this refined embedding. In our implementation, we leveraged the Informer model \cite{informer} for the encoder-decoder, primarily due to its efficiency in long-sequence time-series forecasting. Informer, a transformer-based approach, overcomes the limitations of traditional Transformers through mechanisms such as ProbSparse self-attention, self-attention distilling, and a generative-style decoder. These innovations enhance time and memory efficiency while preserving high predictive accuracy. However, our framework remains model-agnostic and can accommodate various forecasting architectures. For the transfer component, we introduced the use of fully connected layers, graph neural network and attention over locations which we will give more details in Section~\ref{sec:transformcomp}.

\subsection{Retrieval}
\label{chapter5-retrieval}
To augment the zero-shot target point with relevant data, the first step is to retrieve the most similar points to the target. This is achieved by defining a distance function over the points and selecting the closest $k$ stations based on this distance. Additionally, we developed a resolution-aware retrieval mechanism that identifies different sets of points for each frequency band, ensuring that the retrieved data is tailored to the specific resolution requirements of the model.

\subsubsection{Closest Distance Retrieval}

In this method, we select the $k$ closest stations to the target based on a defined distance function. The first function we used is the Haversine distance, which calculates proximity based on latitude and longitude. Additionally, we experimented using a distance function based on the characteristics of the location. By using these characteristics, we can identify points that are not only geographically close but also similar based on their attributes. One way to achieve this is by calculating the Euclidean distance between the Satclip\cite{klemmer2023satclip} location embeddings of the stations. More details about the location embedding techniques will be provided in Section~\ref{sec:location_embedding}.
\subsubsection{Resolution Aware Retrieval}

\begin{figure*}[t]
    \centering
    \includegraphics[width=0.63\textwidth]{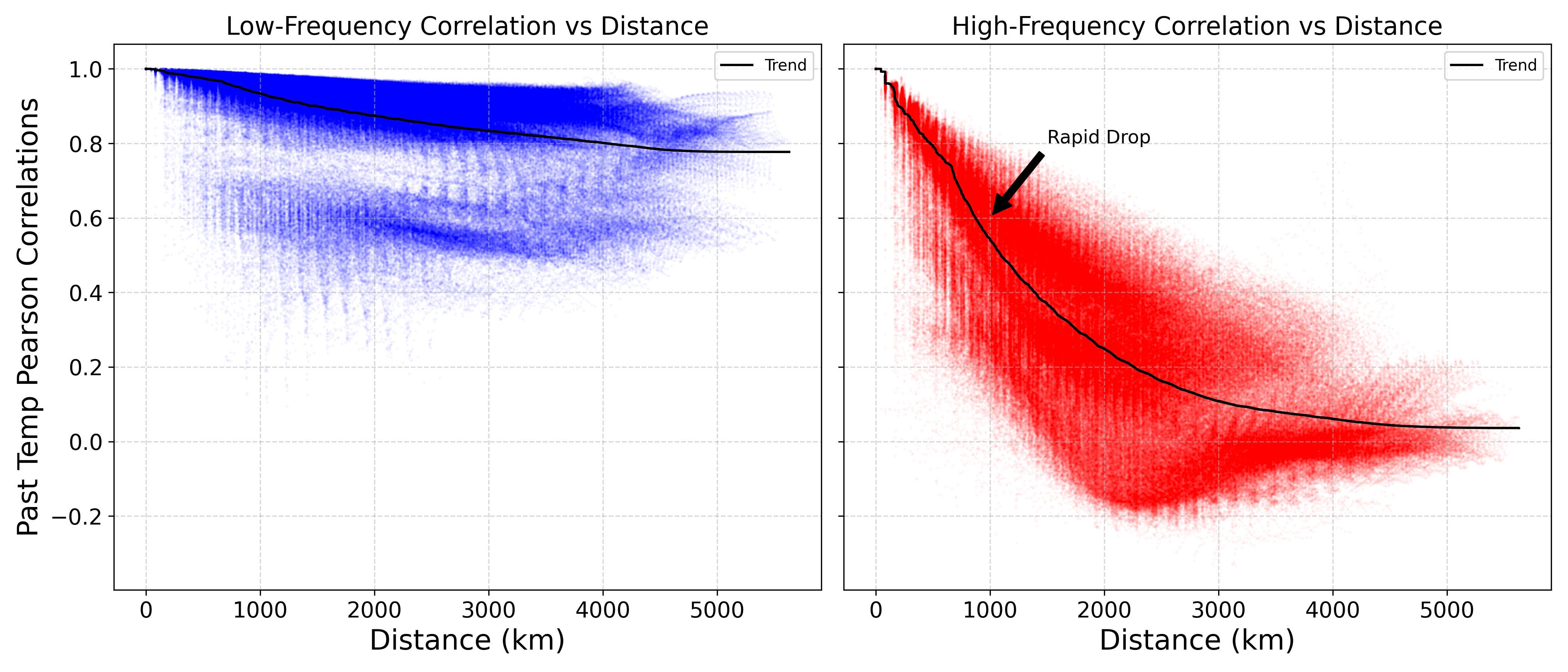}
    \vspace{-1em}
    \caption{Correlation between historical temperature data and distance for a subset of point pairs across the U.S. At low frequencies (left), correlation remains high over long distances, while at high frequencies (right), it declines more rapidly.}
    \label{fig:corr_dist}
    \vspace{-0.5em}
\end{figure*}

In many spatiotemporal time series such as climate, the patterns affect surrounding areas in a way that closer locations have a greater influence on the fast-changing, higher-frequency components, while more distant locations affect slower-changing, lower-frequency components. As illustrated in Figure~\ref{fig:corr_dist}, this is evident in the correlation of points over varying distances for different frequencies. Based on this observation, we developed a model that retrieves different sets of reference points depending on the frequency of the component. 

To implement this, we first decompose the context at the target location using wavelet decomposition: 
$$\mathcal{W}(\mathcal{X}_t(st_{tar})) = \{\mathcal{X}^{f_1}_t(st_{tar}), \mathcal{X}^{f_2}_t(st_{tar}), \mathcal{X}^{f_3}_t(st_{tar}), ...\}$$ 
where $\mathcal{W}$ represents the wavelet decomposition.
The final model, incorporating this resolution-aware retrieval, is defined as: 
$$\hat{\mathcal{Y}}_t(st_{tar}) = \mathcal{W}^{-1}(\{\phi(\mathcal{X}^{f_1}_t(\psi_{f_1}(st_{tar}))), \phi(\mathcal{X}^{f_2}_t(\psi_{f_2}(st_{tar}))), ...\})$$ 
where $\psi_{f_i}(\text{st}_{tar})$ is the set of points retrieved for the target point $\text{st}_{tar}$ at frequency $f_i$, and $\mathcal{W}^{-1}$ is the inverse wavelet transform.

We enforce the constraint: $|\psi_{f_i}(st_{tar})| > |\psi_{f_j}(st_{tar})|,\; if \; f_i < f_j$. In other words, for lower frequencies of data we retrieve a larger set of relevant points.

This ensures that for lower frequencies, where the signal varies more gradually, the model retrieves a larger set of reference points. Additionally, because the wavelet decomposition captures increasingly coarser features at lower frequencies, we also have: 
$$|\mathcal{X}^{f_i}_t| < |\mathcal{X}^{f_j}_t|,\; if \; f_i < f_j$$ 
Meaning that less past context is available for lower frequencies. To illustrate this, consider the example in Figure~\ref{fig:GNN_trans_model}, where only 48 hours of past data is available for the target. If the high-frequency signals are sampled hourly, the model has access to 48 past data points for context. For a moderate-frequency signal, such as daily data, only two past points fall within this window. For a low-frequency signal, such as monthly data, no past context is available within the 48-hour period for the model to adjust to the target.

This design allows the model to rely on a broader context for lower frequencies, improving prediction accuracy where the available data is sparser.

\subsection{Location Embedding}
\label{sec:location_embedding}
We use different strategies for representing a location as a vector embedding which we represent as $\ell(st_i)$. The idea is given the latitude and longitude of the location we want to automatically get a set of features that represent the relevant characteristics of that specific location.

\textbf{Learned Location Embedding:}
We use a fully connected layer which gets latitude, longitude and elevation of the location to learn an embedding for that location, this will be trained with the rest of the model. In this way we learn an embedding on the location that is useful for the rest of the model.

\textbf{Satclip:}We used pre-trained location embedding models that are designed to capture valuable information for a wide range of tasks. One such model is Satclip\cite{klemmer2023satclip}. Satclip learns meaningful characteristics of locations by matching satellite images with their respective coordinates, employing a contrastive location-image pretraining objective. 


\subsection{Transfer Component}
\label{sec:transformcomp}

In our model architecture, we incorporated a transfer component, denoted as $\delta$, which plays a pivotal role in transferring embeddings from source stations to target stations. This component leverages information about both the source and target locations, as well as the embeddings of the source stations. The process can be mathematically described as follows:
$$ E_i'(st_{tar}) = \delta(E(\mathcal{X}_t(st_i)), \ell(st_i), \ell(st_{tar}))$$ 
Here, $E(\mathcal{X}_t(st_i))$ denotes the encoder embedding of the source station $st_i$. The output, 
$E_i'(st_{tar})$, represents the approximated encoder embedding for the target station $st_{tar}$, as influenced by the source station $st_{i}$.

To implement this transfer component, we propose three architectures: a fully connected network, a graph neural network and attention over location embeddings. These architectures enable the model to effectively capture spatial relationships and adaptively transfer information between stations, enhancing the performance of the overall system.
Due to limited space, we explain the location attention module below; other transfer components that we used are summarized here. The FC Transfer module uses a fully connected neural network to transform embeddings from source stations, computing a weighted average to estimate the target station's embedding. The GNN Transfer module represents retrieved points as nodes in a graph with distance-based edge weights, using a Graph Convolutional Network to generate updated embeddings for forecasting.

\textbf{Location Attention as Transfer Component.}
\begin{figure*}[t]
    \centering
    \includegraphics[width=0.63\textwidth]{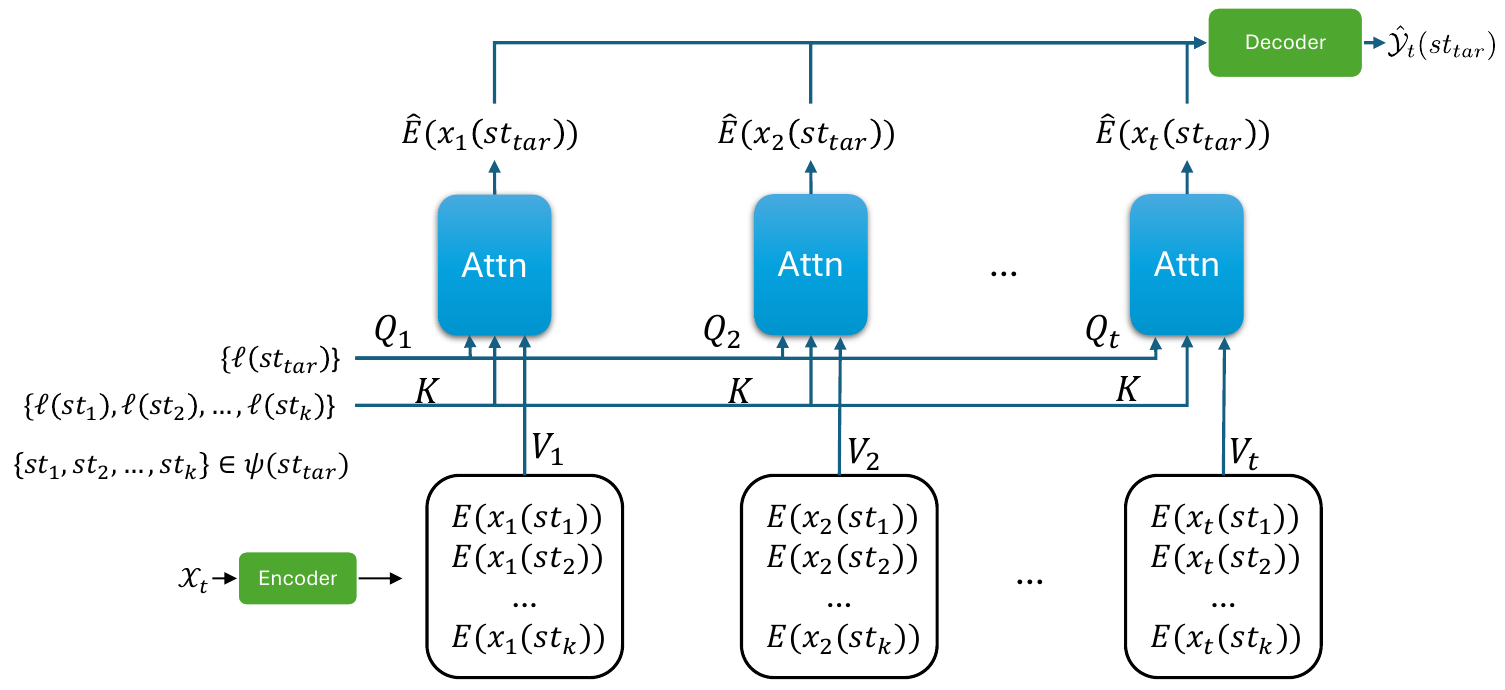}
    \vspace{-1em}
    \caption{Retrieval augmented forecasting model that uses attention over location embeddings as the transfer module.
    }
    \label{fig:RAG_locattn}
    \vspace{-0.5em}
\end{figure*}
As illustrated in Figure~\ref{fig:RAG_locattn}, our approach trains an attention model that uses the target location’s embedding as a query and the location embeddings of retrieved neighboring points as keys. To account for varying temporal dynamics, we employ a linear projection that extends the target query embedding across multiple timesteps. This design enables the model to attend to different retrieved points at different time steps, thereby capturing important temporal and spatial dependencies.

Once the attention weights are computed, the model aggregates the embeddings of the most relevant retrieved locations into a single representation for the target location. This aggregated embedding is then passed to the decoder, which generates the final forecast. By dynamically adjusting which retrieved points receive the most attention at each timestep, the model is better able to exploit local context.

\subsection{Probabilistic Forecasting}

Forecasting not just the exact value of a predictand but also the range it is likely to fall within is often more informative and practical. One approach to achieve this is by predicting a probabilistic distribution for the forecasted values. Since the introduced model components are agnostic to the specific forecasting architecture, the encoder-decoder in our framework can be replaced with a probabilistic forecasting model. To this end, we developed a probabilistic version of the Informer forecasting model that predicts the parameters of a Gaussian distribution—mean ($\mu_t$ ) and variance ($\sigma^2$)—at each time step, instead of providing deterministic point forecasts. This probabilistic model seamlessly integrates into our architecture, replacing the original encoder-decoder components.

The probabilistic model is trained using the negative log-likelihood (NLL) loss function, which measures how well the predicted Gaussian distribution fits the observed data. For a single data point, the NLL loss is defined as:
$$\mathcal{L}_{NLL} = \frac{1}{2}\sum_t \big(\log{\sigma^2_t} + \frac{(y_t - \mu_t)^2}{\sigma_t^2}\big)$$
This approach allows the model to quantify uncertainty in its predictions, providing both a central estimate and a measure of confidence at each time step, which is particularly valuable for decision-making in high-stakes scenarios.

\subsection{Training Procedure}
Our training methodology consists of two distinct phases. Initially, we train the model to forecast the data and we use the historical data of all train stations. During this phase, the model parameters are updated to capture the global patterns and relationships within the data. Then we pick each of the train stations as the target station, freezing the model parameters except the transfer module in the retrieval-augmented forecasting model and we train these weights.

\section{Experiments}
\label{chapter5:exp}
In this section, we evaluate the performance of our models against the use of the Informer model as an encoder-decoder without the transfer component, the resolution-aware retrieval approach, as well as deep learning contenders and classical baselines 
. We tried our models on a subset of the global high resolution weather dataset called ERA5\cite{hersbach2020era5}.


\subsection{ERA5 Dataset}


\begin{table*}[t!]
\centering
\begin{tabular}{ccccc}
\hline\hline
\textbf{Model} & \textbf{MSE}$\downarrow$ & \textbf{MAE}$\downarrow$ \\
\hline
Informer & 19.05 & 3.19 \\
Informer + GNN transfer & 18.57 & 3.22 \\
Informer + FC transfer & 17.85 & 3.11 \\
Informer + Loc Attn transfer & \underline{\textbf{17.76}} & \underline{\textbf{3.10}} \\
\hline\hline
\end{tabular}
\caption{Average mean squared error (MSE) and mean absolute error (MAE) in Fahrenheit for zero-shot forecasting of ERA5 Pacific North West data using different transfer modules. The average is calculated on the validation portion of the 10 zero-shot points in our validation set.}
\label{tab:GNNResults}
\end{table*}


\begin{table*}[ht!]
    \centering
    \renewcommand{\arraystretch}{1.2}
    \setlength{\tabcolsep}{10pt}
    \begin{tabular}{ccccc}
         & \textbf{Model} & \textbf{MSE$\downarrow$} & \textbf{MAE$\downarrow$} & \textbf{MAPE(\%)$\downarrow$} \\
    \hline
    \multirow{6}{*}{\textbf{Baselines}} 
      & Last Value       & $80.97$  & $6.50$  & $16.33$ \\
      & Moving Average   & $64.86$  & $6.16$  & $18.70$ \\
      & Persistence      & $43.90$  & $4.93$  & $14.62$ \\
      & SARIMA           & $41.74$  & $4.61$  & $11.84$ \\
      & Seasonal Na\"ive  & $33.40$  & $4.25$  & $12.16$ \\
      & AutoReg  & $24.15$  & $3.56$  & $10.74$ \\
    \hline
    \multirow{5}{*}{\shortstack{\textbf{Large}\\
    \textbf{Foundation}\\
    \textbf{Models}}} 
      & TimesFM          & $29.86$  & $3.98$  & $12.02$ \\
      & Chronos base     & $26.99$  & $3.80$  & $10.92$ \\
      & Chronos large    & $26.60$  & $3.78$  & $10.82$ \\
      & Chronos Bolt     & $26.37$  & $3.69$  & $10.28$ \\
      & Chronos Bolt FT  & $22.58$  & $3.44$  & $9.94$ \\
    \hline
    \multirow{4}{*}{\textbf{Ours}} 
      & Informer             & $19.05$  & $3.19$  & $9.80$ \\
      & Informer + Dec       & $19.76$  & $3.31$  & $10.11$ \\
      & Informer + Ret       & $17.95$  & \underline{\textbf{3.11}}  & $9.46$ \\
      & Informer + Dec + Ret   & \underline{\textbf{17.79}}  & $3.15$  & \underline{\textbf{9.32}} \\
    \hline
    \end{tabular}
    \captionof{table}{Results of the models averaged over the test set on ERA5 Pacific North West dataset. 
    }
    \label{tab:ERA5Results}
    \vspace{-2em}
\end{table*}

\begin{table}[h]
\centering
\begin{tabular}{lccc}
\hline
\textbf{Model} & \textbf{MAPE (\%)} & \textbf{sMAPE (\%)} & \textbf{MASE} \\
\hline\hline
Chronos Bolt FT & 9.94 & 8.27 & 0.70 \\
TimesFM\_v2     & 11.79 & 9.30 & 0.78 \\
Informer        & 9.80 & 7.71 & 0.65 \\
Informer+Dec+Ret & \underline{\textbf{9.32}} & \underline{\textbf{7.62}} & \underline{\textbf{0.64}} \\
\hline
\end{tabular}
\caption{sMAPE and MASE performance comparison across models.}
\label{tab:sMAPE_MASE}
\end{table}

\begin{table}[t!]
\centering
\begin{tabular}{cccc}
\hline\hline
\textbf{Model} & \textbf{MSE$\downarrow$} & \textbf{MAE$\downarrow$} & \textbf{MAPE}$(\%)\downarrow$ \\
\hline
HRRR          & $36.91$ & $4.47$  & $12.50$ \\
Chronos       & $16.70$  & $3.01$  & $9.18$ \\
Informer + Ret         & \underline{\textbf{10.82}}  & \underline{\textbf{2.42}}  & \underline{\textbf{7.68}} \\
\hline
\end{tabular}
\caption{Average MSE, MAE and MAPE on zero shot forecasting of test stations in ERA5 Pacific North West dataset comparing with HRRR, a numerical weather prediction (NWP) model. The results are averaged over 24 hours of forecast horizon.}
\label{tab:HRRRresults}
\vspace{-2.5em}
\end{table}

\begin{figure*}[t!]
    \centering
    \includegraphics[width=0.71\textwidth]{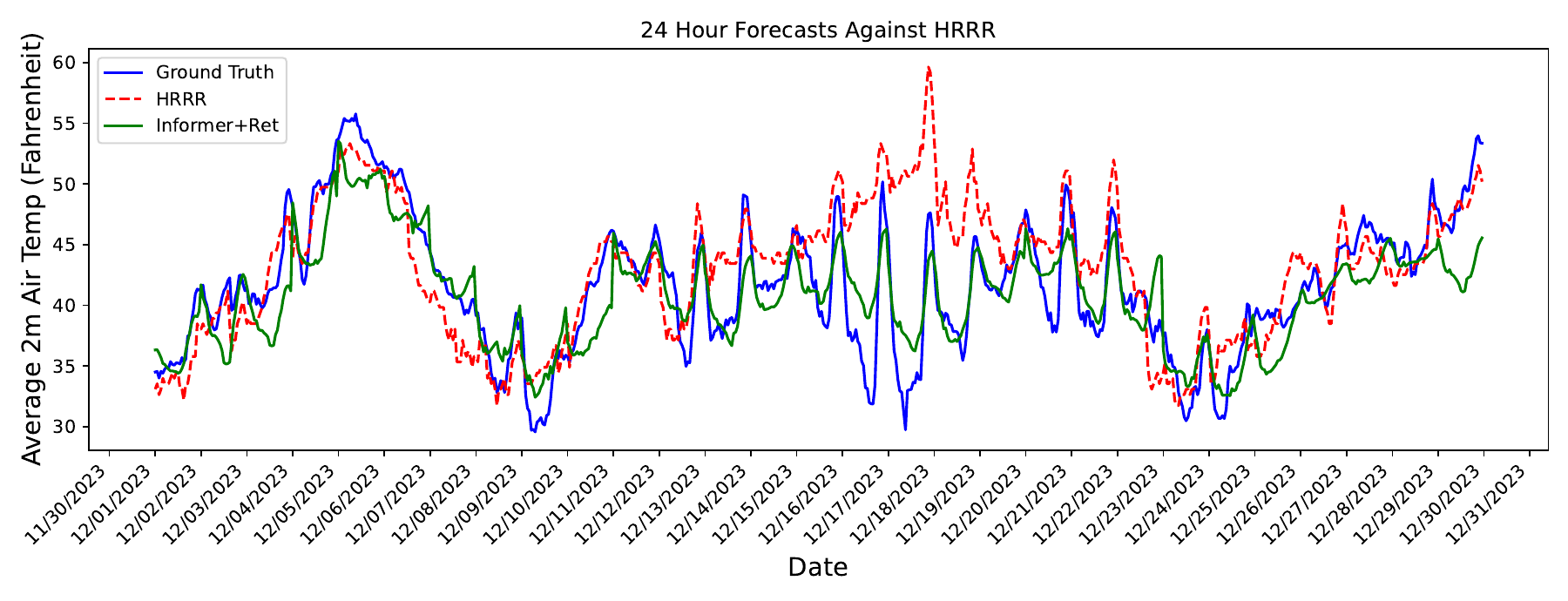}
    \vspace{-1em}
    \caption{Forecasts of our full model agianst those of HRRR and ground truth. The forecasts are ran at every midnight for 24 hours for both models.}
    \vspace{-2pt}
    \label{fig:PredvsGround}
\end{figure*}

\begin{figure*}[t]
    \centering
    \includegraphics[width=0.4\textwidth]{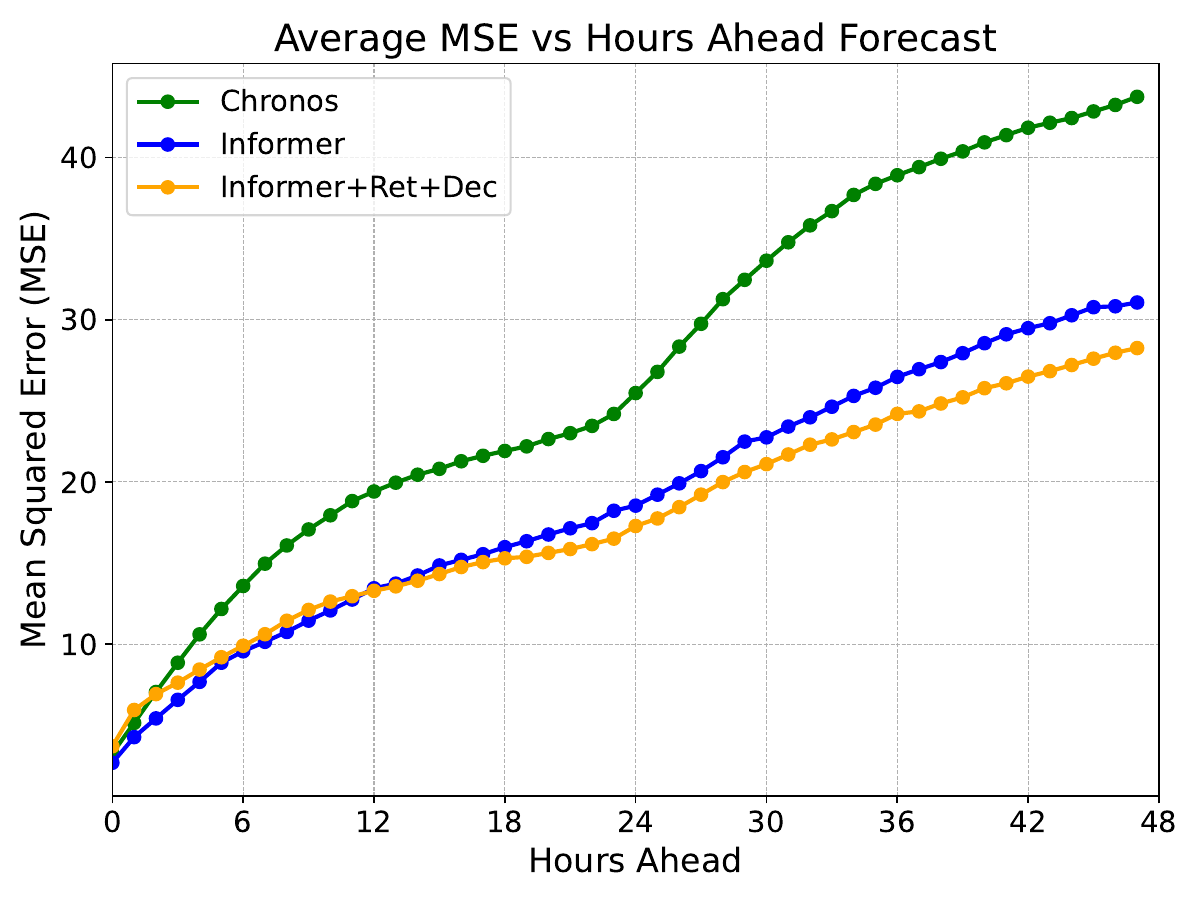}
    \includegraphics[width=0.4\textwidth]{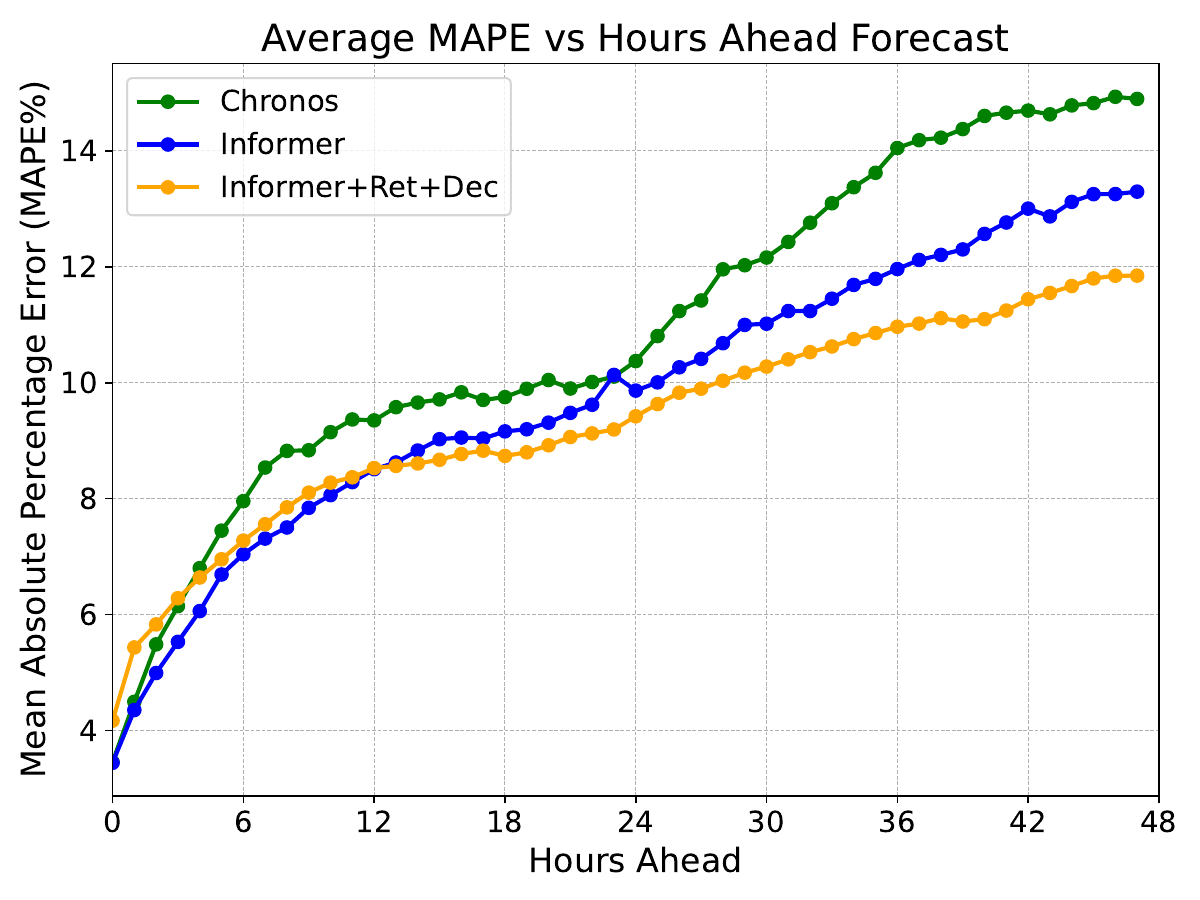}
    \vspace{-1em}
    \caption{MSE and MAPE for test data for each hours of forecast for three models of Chronos base, Informer model trained on the training stations and the full model with resolution-aware retrieval (Informer + Ret + Dec).} 
    \label{fig:MSE_48hrs}
    \vspace{-1em}
\end{figure*}
\begin{figure*}[t]
    \centering
    \includegraphics[width=0.45\textwidth]{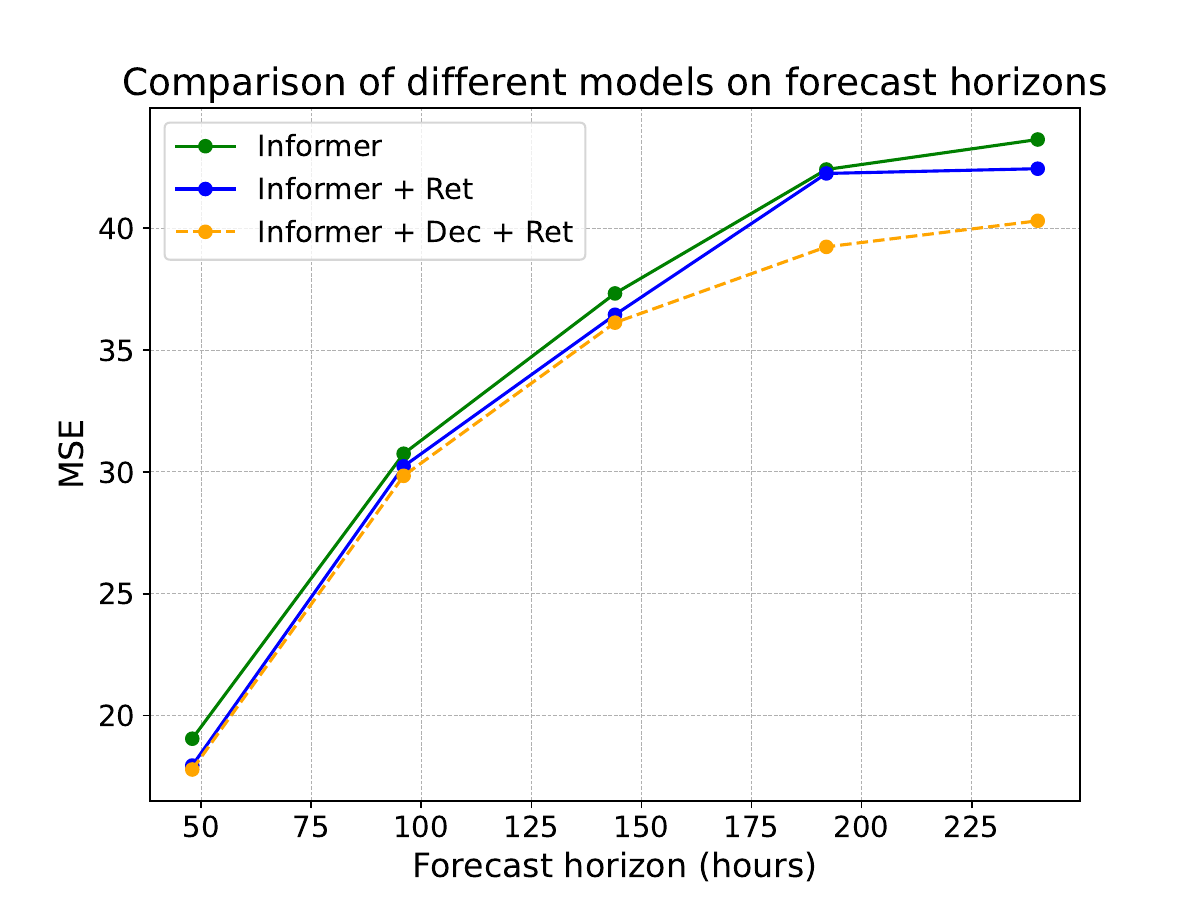}
    \includegraphics[width=0.45\textwidth]{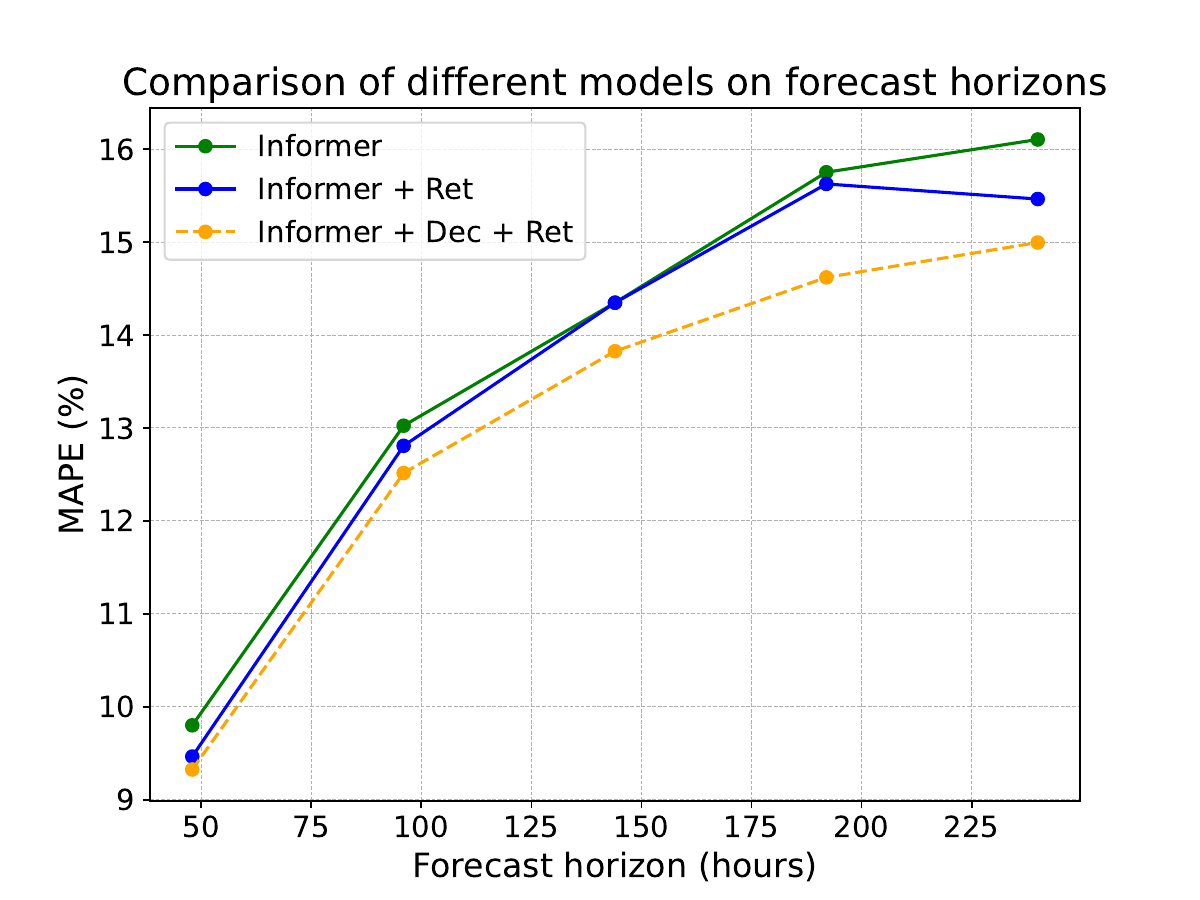}
    \vspace{-1em}
    \caption{MSE, and MAPE for test data for different forecast horizons. The results for our trained models, Informer, Informer with retrieval (Informer + Ret) and full model with resolution-aware retrieval (Informer + Dec + Ret) are shown.} 
    \vspace{-0.5em}
    \label{fig:Forecast_Horizons}
\end{figure*}


\begin{figure*}[t]
    \centering
    \includegraphics[width=0.4\textwidth]{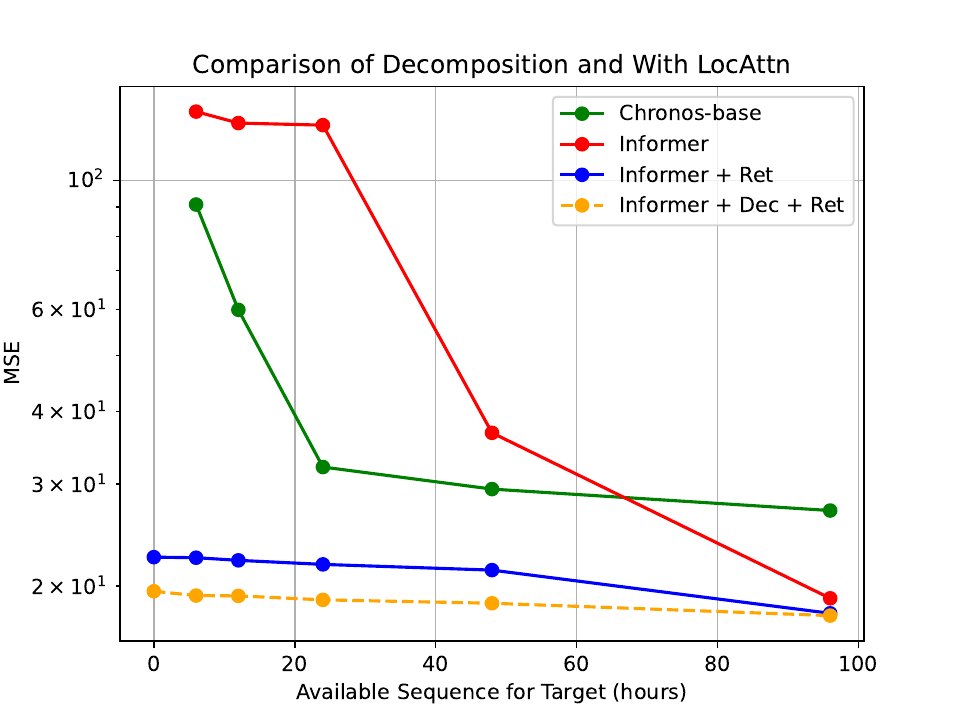}
    \includegraphics[width=0.4\textwidth]{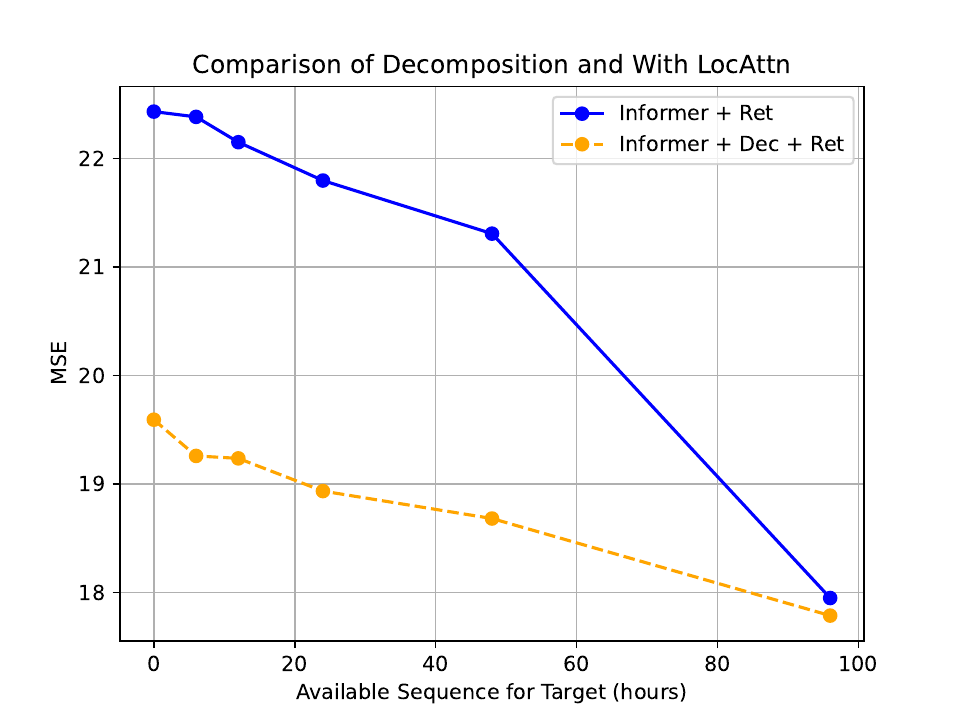}
    \vspace{-1em}
    \caption{MSE vs. different sizes of sequence available for the target, Left: The results with Chronos, and Informer without retrieval, note that due to non-retrieval models' high errors in this scenario the y-axis is given as log-scale, Right: Results with our trained Informer model with retrieval that uses location attention transfer module (Informer + Ret) and full model with resolution-aware retrieval(Informer + Dec + Ret).} 
    \label{fig:Diff_SeqSize}
\end{figure*}

We utilized the global ERA5 dataset \cite{hersbach2020era5}, provided by the European Centre for Medium-Range Weather Forecasts (ECMWF). Initially, we created a subset of this dataset focusing on the US Pacific Northwest to experiment with our models. Subsequently, we trained the best-performing models using data from the entire United States, enabling them to serve as foundation models for any location within the country. The dataset includes five climate variables for each data point, spanning from January 1, 2020, to January 31, 2023.

For temporal splitting, we divided the data as follows: the first 70\% from training stations was used for model training, 10\% for validation (from left-out validation stations), and the final 20\% from zero-shot locations for testing. The target variable is the 2-meter above-ground temperature (t2m), measured in degrees Kelvin which we convert to degrees Fahrenheit.  Input features include historical values of the eastward and northward wind components at 10 meters above ground, temperature and dew point temperature at 2 meters above ground, and surface pressure.

Using data from the previous 96 hours ($L_x = 96$), at each hour we generated forecasts for the subsequent 48 hours ($L_y = 48$) for the 2-meters above ground temperature. Errors were calculated and reported for each hour of the forecast horizon.

\subsubsection{Pacific North West}
 We initially restricted the dataset to 320 points within the rectangular area bounded by ($48.75^\circ N, 124^\circ W$) and ($44^\circ N, 120.24^\circ W$).
For our experiments, we randomly designated 10 points as zero-shot test locations and another 10 points for validation, training our models on the remaining 300 points.

In Table~\ref{tab:GNNResults}, first we compare the performance of the Informer model with three different transfer modules: FC transfer, GNN transfer, and Location Attention (Loc Attn) transfer. For all models, 50 retrieved points were used. Our results demonstrate that the location attention transfer module consistently outperforms both FC and GNN transfer on the validation set of this dataset.

Table~\ref{tab:ERA5Results} presents the mean squared error (MSE), mean absolute error (MAE), and mean absolute percentage error (MAPE) results of our model, averaged over the 10 left-out test stations, alongside comparisons with several baseline models. Notably, our model outperforms these baselines, as well as large foundation time series models such as Chronos \cite{ansari2024chronos} and TimesFM \cite{das2023decoder}. To establish a stronger baseline, we fine-tuned the Chronos Bolt model on the training portion of this dataset, referred to as Chronos Bolt FT. Our model outperforms these large foundation models due to its ability to adapt quickly to the target location, whereas the baselines and large foundation models struggle with the limited context available for the target location. We also report the Symmetric Mean Absolute Percentage Error (sMAPE) and Mean Absolute Scaled Error (MASE)—which compares model performance against a na\"ive seasonal forecast—for Chronos Bolt FT, TimesFM\_v2 \cite{das2023decoder}, and our full model in Table~\ref{tab:sMAPE_MASE}.

In scenarios like this, high-resolution numerical weather prediction models are commonly used, with High-Resolution Rapid Refresh (HRRR) \cite{benjamin2016north} being one of the most notable. Since HRRR does not provide forecasts for 48 hours ahead, we limited the forecasting horizon to 24 hours to enable a direct comparison. The results, shown in Table~\ref{tab:HRRRresults}, demonstrate that our models significantly outperform HRRR on the test set. 
Figure~\ref{fig:PredvsGround} shows the forecasts of HRRR vs. our best model for December 2023 for a random zero-shot location. Note that while numerical weather prediction models such as HRRR are considered the standard for weather forecasting and are highly accurate in general, they have been shown to be less reliable for microclimate predictions. This is because they attempt to model weather patterns on a global scale at relatively lower resolutions, making it challenging to capture localized variations effectively. Prior research has also highlighted these limitations \cite{kumar2021micro}.

In Figure~\ref{fig:MSE_48hrs}, we provide MSE, and MAPE results across different forecasting horizons, comparing Chronos, Informer, and our full model. While Informer and Chronos demonstrate better performance for short-term forecasts (a few hours ahead), our model outperforms them for longer forecasting horizons (12+ hours ahead). This improvement is attributed to our retrieval and decomposition processes, which enhance the prediction of lower-frequency components—critical for accurate long-term forecasts.

Figure~\ref{fig:Forecast_Horizons} further illustrates forecasting errors for three setups: the Informer model alone, Informer with retrieval (Informer + Ret), and Informer with resolution-aware retrieval (Informer + Dec + Ret). As the forecasting horizon increases, the model with resolution-aware retrieval consistently outperforms the other two configurations. This advantage stems from its ability to better capture lower-frequency and longer-term trends in the data.

Figure~\ref{fig:Diff_SeqSize} illustrates the impact of reduced context data availability for the target station. The resolution-aware retrieval model significantly outperforms the basic retrieval-augmented model and baseline models like Chronos, which exhibit substantially higher errors in this scenario. Notably, Chronos and other non-retrieval models struggle to adapt to new locations with limited past context—highlighted by the log-scale on the y-axis in the left figure—whereas our resolution-aware approach effectively leverages the available data.

On the right side of Figure~\ref{fig:Diff_SeqSize}, we compare two retrieval methods: one using resolution-aware retrieval (Informer+Dec+Ret) and the other using a standard retrieval-augmented forecasting model without resolution-aware retrieval (Informer+Ret). The results clearly show that the resolution-aware retrieval model outperforms the standard retrieval model, as it retrieves relevant points for different frequencies, making it more robust to limited context availability for the target point.


\subsubsection{Probabilistic Forecasting}
\begin{table}[t!]
    \centering
    \begin{tabular}{c c c c c c}
    Model & NLL$\downarrow$ & CRPS$\downarrow$ & \makecell{Coverage \\ (0.95)} & MAE$\downarrow$\\ \hline
    Chronos Bolt & $3.61$ & $3.64$ & $0.84$ & $3.69$ \\
    ProbInformer & $2.75$ & $2.26$ & \underline{\textbf{0.93}} & $3.17$ \\
    ProbInformer+Ret & \underline{\textbf{2.72}} & \underline{\textbf{2.20}} & \underline{\textbf{0.93}} & \underline{\textbf{3.08}} \\
    \hline
    \end{tabular}
\caption{Results of the probabilistic versions of the model. We show negative log-likelihood (NLL), Countinuous Ranked Probability Score (CRPS), 0.95 coverage, and MAE results of our test set are shown.} 
\label{tab:ProbResults}
\vspace{-2em}
\end{table}


\begin{figure*}[ht!]
    \centering
    \includegraphics[width=0.48\textwidth]{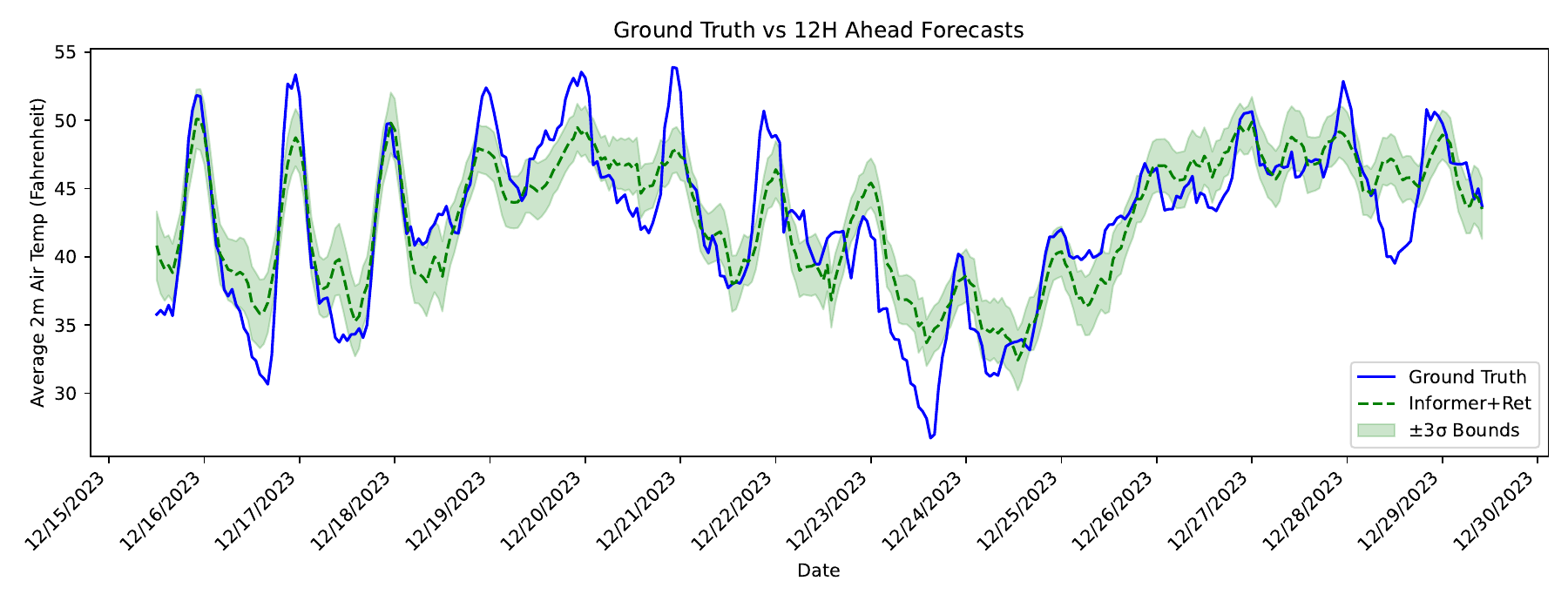}
    \includegraphics[width=0.48\textwidth]{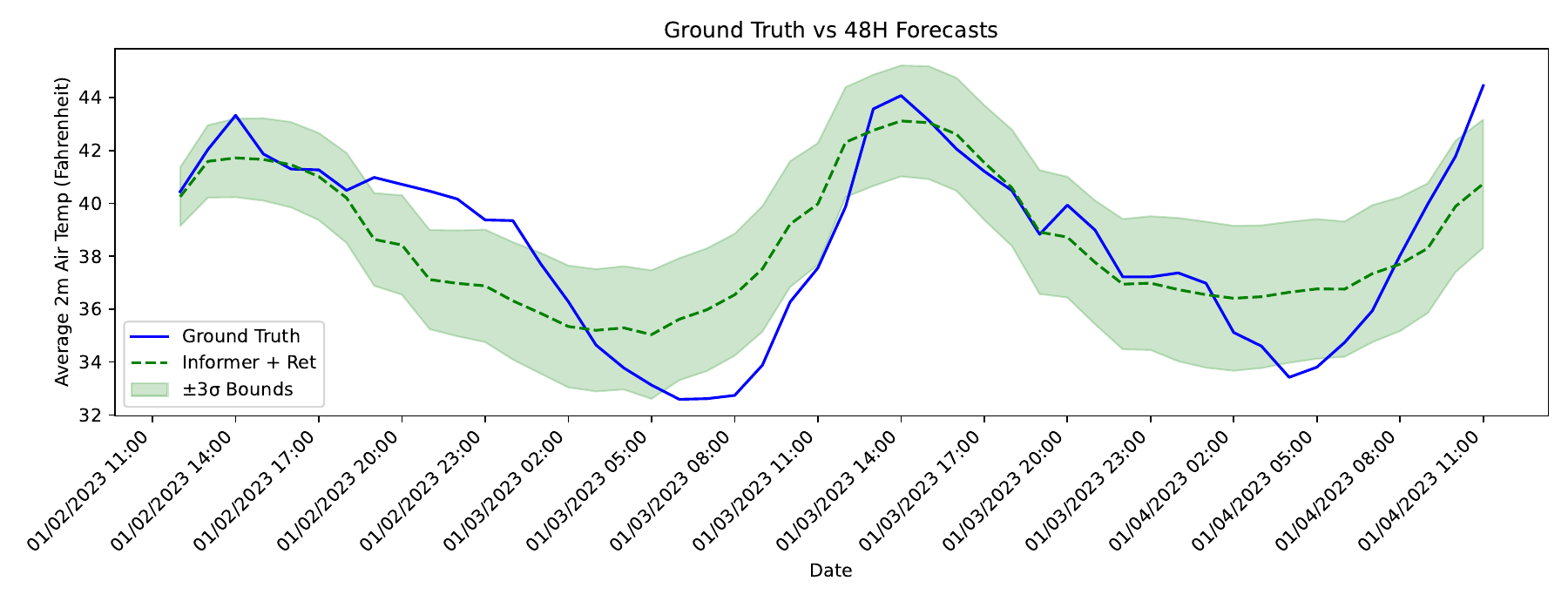}
    \caption{Left: 12 hours ahead forecasting of temperature for our model vs. the ground truth for last two weeks in 2023, Right: The 48 hours prediction of our model for January 2nd of 2023. 3 $\sigma$ bound is shown for both plots.} 
    \label{fig:ProbPlots}
\end{figure*}

In many weather and climate prediction scenarios, it is crucial to consider not only the deterministic forecasted values but also the range of possible values. In this study, we evaluate the probabilistic model we developed, ProbInformer, by analyzing its performance both with and without the retrieval component.

To evaluate our models, we used several metrics: Negative Log-Likelihood (NLL), Continuous Ranked Probability Score (CRPS), Coverage, and Mean Absolute Error (MAE) of the predicted mean. CRPS assesses the accuracy of probabilistic forecasts by comparing the predicted cumulative distribution function (CDF) against the observed values, serving as a probabilistic generalization of MAE. Coverage quantifies how frequently the true observations fall within the predicted intervals, providing a measure of the calibration of these intervals.

In Table~\ref{tab:ProbResults}, we present the average NLL, CRPS, $\text{Coverage}_{0.95}$, and MAE results for our Gaussian model, ProbInformer, and its version incorporating retrieval (ProbInformer+Ret). We compare these results with the Chronos Bolt model. For Chronos Bolt, which provides quantiles and the mean forecast, we estimated the standard deviation of a Gaussian distribution for each time step using the provided quantiles. Our results show that ProbInformer and ProbInformer+Ret achieve better NLL and CRPS scores. Additionally, our models are well-calibrated, with Coverage metrics close to their respective targets, although they slightly underestimate uncertainty.

In Figure~\ref{fig:ProbPlots}, we illustrate 12-hour-ahead forecasts and 48-hour forecasts for our model at the location $44.0^\circ \text{N}, 122.75^\circ \text{W}$. The left plot displays forecasts for the last two weeks of 2023, while the right plot shows the full 48-hour forecast initialized on January 2, 2023. In both plots, the $3 \sigma$ bounds are also shown. Notably, the model predicts greater uncertainty when the forecasts exhibit higher errors. 

\subsubsection{Wind Speed Prediction}
\label{app:wind_speed_pred}

To compute the wind speed from the u-component (\( u \)) and v-component (\( v \)) of the wind, we use the following formula:
$$\text{Wind Speed} = \sqrt{u^2 + v^2}$$
Both components are initially given in meters per second (m/s) and we convert them to miles per hour (mph). 
This transformation allows for a direct comparison of wind speed predictions across different models. The performance of each model is evaluated using Mean Squared Error (MSE) and Mean Absolute Error (MAE). The results are summarized in Table~\ref{tab:HRRRresults_Windspeed}.

\begin{table}[ht!]
\centering
\begin{tabular}{ccc}
\hline\hline
\textbf{Model} & \textbf{MSE$\downarrow$} & \textbf{MAE$\downarrow$} \\
\hline
HRRR & 12.28 & 2.56 \\
Informer & 5.01 & 1.50 \\
Informer + Ret & \underline{\textbf{4.88}} & \underline{\textbf{1.48}} \\
\hline
\end{tabular}
\caption{Average MSE, and MAE on zero shot wind speed forecasting of test stations in ERA5 Pacific North West dataset comparing with HRRR, a numerical weather prediction (NWP) model. The results are in miles per hour (MPH) and are averaged over 24 hours of forecast horizon.}
\label{tab:HRRRresults_Windspeed}

\end{table}

\subsubsection{Frozen Events Prediction}
While MAE and MSE provide general accuracy measures, they do not directly reflect the model's usefulness in critical decision-making applications such as predicting freeze events (temperatures below 32°F). To assess the models' effectiveness in this context, we evaluate the F1-score for freeze event detection over a 24-hour forecasting horizon, comparing our models against HRRR and Chronos. The results are presented in Table~\ref{tab:freeze_results}.

\begin{table}[ht!]
\centering 
\begin{tabular}{c c} 
\hline\hline
\textbf{Model} & \textbf{F1-score} $\uparrow$ \\\hline 
HRRR & 0.613 \\
Chronos & 0.627 \\ 
Informer+Ret & 0.688 \\ 
Informer+Dec+Ret & \textbf{0.736} \\  
\end{tabular} 
\caption{F1-scores for freeze event detection in 24-hour forecasts, comparing different models.} 
\label{tab:freeze_results} 
\end{table}

Among all models, our full model (Informer+Dec+Ret) achieves the highest F1-score (0.736), indicating the best balance between precision and recall for freeze event detection. Retrieval Augmented Forecasting (Model+Ret) performs (0.688) better than Chronos (0.627), benefiting from retrieval-based adaptation.

These results highlight the advantages of retrieval-augmented and resolution-aware forecasting models, which effectively capture localized climate variations and enhance predictive performance in critical scenarios.

\subsubsection{Entire US}
Finally, we trained our best model using 23,067 regularly spaced points across the United States, randomly leaving out 100 points, of which 50 are used as zero-shot test locations. Our best models achieved an average MAE of $2.97$ and MAPE of $12.47\%$ for 24-hour predictions of 2-meter above-ground temperature in degrees Fahrenheit over the test data period. These models serve as foundation models, capable of performing zero-shot microclimate forecasts for any location in the United States without requiring prior data. They offer a robust framework for zero-shot and cold-start microclimate forecasting.


\section{Conclusion}
\sloppy
In this work, we introduced a novel approach that integrates resolution-aware retrieval and retrieval-augmented forecasting to address the challenge of generating accurate predictions for locations with sparse or no training data. By extrapolating embeddings from well-trained source areas to target locations, our model enables precise zero-shot forecasting. Leveraging personalization through retrieval-augmented modeling and multi-resolution learning, the method dynamically identifies and retrieves relevant data points to enhance forecasting performance.

We evaluated our approach extensively on 
real-world datasets, focusing on temperature prediction in microclimate setting. The results demonstrate that our model outperforms HRRR, baseline methods, and other deep learning models in generating forecasts for previously unmonitored locations. This work highlights the effectiveness of retrieval-augmented and resolution-aware strategies in advancing the accuracy and generalizability of forecasting systems in zero-shot scenarios.

While we evaluated the model primarily for temperature prediction and included some results for wind speed prediction in microclimate setting, other microclimate parameters such as solar radiation can be easily calculated. Furthermore, our model's design readily extends to other spatiotemporal applications such as traffic flow analysis, epidemic modeling, energy demand forecasting, and environmental monitoring. Additionally, future research could explore extending the framework to non-spatiotemporal domains. If the relationship between subjects' covariates and temporal resolutions can be identified, the model might be adapted for applications such as demand forecasting or other domains requiring structured retrieval. A promising future direction is to learn these relationships dynamically, enabling the model to optimize the retrieval function for each resolution and further refine its predictive capabilities.


\bibliographystyle{ACM-Reference-Format}
\bibliography{references}

\clearpage

\end{document}